\documentclass[runningheads]{llncs}
\usepackage[T1]{fontenc}
\usepackage{amssymb}
\usepackage{enumitem}
\usepackage[usenames]{color}

\usepackage{graphicx}
\begin{document}
\title{VideoRun2D: Cost-Effective Markerless \\Motion Capture for Sprint Biomechanics}
\titlerunning{VideoRun2D: Cost-Effective Markerless Motion Capture}
%

\author{
Gonzalo Garrido-Lopez\inst{1} \and
Luis F. Gomez\inst{2} \and
Julian Fierrez\inst{2}\and
Aythami Morales\inst{2} \and
Ruben Tolosana\inst{2} \and
Javier Rueda\inst{1} \and
Enrique Navarro\inst{1} 
}

\authorrunning{G. Garrido-Lopez and L.F. Gomez and J. Fierrez et al.}
%


\vspace{.5cm}

\institute{Sport Biomechanics Laboratory, Faculty of Physical Activity and Sports Sciences INEF, Universidad Politécnica de Madrid, Madrid, Spain. \\
\email{\{gonzalo.garrido.lopez, javier.ruedao, enrique.navarro\}@upm.es} \vspace{.5cm}
\and
BiDA Lab, Universidad Autonoma de Madrid, Madrid, Spain.\\
\email{\{luisf.gomez, julian.fierrez, aythami.morales, ruben.tolosana\}@uam.es}
}
\maketitle              
\begin{abstract}
Sprinting is a determinant ability, especially in team sports. The kinematics of the sprint have been studied in the past using different methods specially developed considering human biomechanics and, among those methods, markerless systems stand out as very cost-effective. On the other hand, we have now multiple general methods for pixel and body tracking based on recent machine learning breakthroughs with excellent performance in body tracking, but these excellent trackers do not generally consider realistic human biomechanics (like the ones considered in support for professional sports or injury rehabilitation). This investigation first adapts two of these general trackers (MoveNet and CoTracker) for realistic biomechanical analysis and then evaluate them in comparison to manual tracking (with key points manually marked using the software Kinovea). 
Our best resulting markerless body tracker particularly adapted for sprint biomechanics is termed VideoRun2D. The experimental development and assessment of VideoRun2D is reported on forty  sprints recorded with a video camera from 5 different subjects, focusing our analysis in 3 key angles in sprint biomechanics: inclination of the trunk, flex extension of the hip and the knee. The CoTracker method showed huge differences compared to the manual labeling approach. However, the angle curves were correctly estimated by the MoveNet method, finding errors between 3.2° and 5.5°. 
In conclusion, our proposed VideoRun2D based on MoveNet core seems to be a helpful tool for evaluating sprint kinematics in some scenarios. On the other hand, the observed precision of this first version of VideoRun2D as a markerless sprint analysis system may not be yet enough for highly demanding applications. Future research lines towards that purpose are also discussed at the end: better tracking post-processing and user- and time-dependent adaptation.

\end{abstract}

\section{Introduction}
\setcounter{footnote}{0} 

Recent advances in human tracking technology, driven by innovations in computer vision, machine learning, and sensor fusion, have significantly enhanced the accuracy, scalability, and versatility of tracking systems. Human tracking technology has the potential to transform sports by enabling precise performance analysis, injury prevention, and personalized training, providing athletes and coaches with real-time insights to optimize strategies and enhance overall performance.

Sprinting is an important ability used in many sports~\cite{morin2011technical}, and its analysis is key for both performance improvement in sports and injury prevention. 
In the past, observational analysis was used to evaluate different human movement patterns~\cite{yang2024improving}. This method lacks objective measures and consistency between observers and requires clinical expertise~\cite{hii2023automated,yang2024improving}. Manual marking systems were created to obtain objective information, calculating time variables and joint angles during running~\cite{bissas2022kinematic,hanley2022biomechanics}. This technique is still used for evaluating sprinting kinematics, especially in competitions~\cite{martinez2021electromyography}. These methods are cheap and easy to use, but they require a large quantity of time, and they are applied by human operators, so errors could appear when processing a sprint video~\cite{cronin2024feasibility}.

Other methods, such as Inertial Measurement Units (IMUs), have been used to estimate kinematic variables but those devices have some methodological problems~\cite{acien2020sensors,bastiaansen2020inertial,lin2023validity,nazarahari2022foot}.
To increase accuracy, marker-based systems were created~\cite{ota2021verification,yang2024improving}. These systems are widely used for gait and sprinting analysis because they are precise, although they are expensive and time-consuming, and all the measures are taken in a controlled environment by specialized personnel~\cite{paula2023gait,viswakumar2022development,yang2024improving}.

On the other hand, smartphone apps for analyzing sprint variables have appeared for coaches who do not usually have access to other more specialized and expensive methods~\cite{romero2017sprint}. These apps estimate variables such as theoretical maximal force, velocity during the sprint, etc., but do not estimate kinematic variables of joint angles~\cite{romero2017sprint}. This gap is being filled by markerless systems, which are emerging as a potential solution due to their cost-effectiveness ratio and faster results. However, they still require high computational resources and human supervision, and the accuracy remains unclear~\cite{stenum2021two,yang2024improving}, specially for biomechanical analysis. One of the first markerless systems used for running analysis was Kinect, but some studies have demonstrated that this system failed to estimate joint angles properly under different conditions~\cite{ota2021verification,takeichi2018mobile,viswakumar2022development}.

Video data is used as input to markerless motion capture algorithms that usually employ some kind of neural network to detect precisely human joints using only a smartphone camera~\cite{acien2020sensors,viswakumar2022development,yang2024improving}. These techniques use pose estimation techniques, computer vision, and machine learning algorithms to track the human movement of the body limbs and joints over time~\cite{hii2023automated}, 
with excellent results. On the downside, these excellent body trackers do not generally consider realistic human biomechanics, as the main objective in their development (trained usually with automatic machine learning procedures) is the minimization of some kind of visual errors through specific loss functions~\cite{2022_PR_SetMargin_Morales,2022_AI_SensitiveLoss_IS} not considering biomechanical restrictions. The visual output of those trackers may look very nice, but the underlying joint angles and limb trajectories may not represent well the reality of the analyzed body skeleton. As highlighted in the literature \cite{2023_SNCS_Human-Centric_Pena}, it is important to look back at training data and procedures in machine learning-based methods in order to properly apply them in a specific setup, e.g., a general tracker for sprint analysis as studied here.

The contributions of the present work are as follows:

\begin{itemize}
    \item We present VideoRun2D, a new markerless biomechanic sprint analysis system based on deep learning models.
    \item We evaluate VideoRun2D under different configurations (e.g., two body tracker cores: MoveNet and CoTracker) versus a manually-marked ground truth (Kinovea) to evaluate the biomechanic errors between the configurations.
    \item We make publicly available in GitHub the new datasets we have generated for our experiments: sprint videos (40 in total) and associated biomechanical features across time (the best obtained with our automatic VideoRun2D and the manually-marked ground truth)\footnote{https://github.com/BiDAlab/VideoRun2D}.
\end{itemize}

We are sure that these contributions will help to move forward current technology for cost-effective sprint analysis and beyond: any other application of general body trackers to concrete type of movements or application fields, e.g., professional sports analysis and improvement, injury analysis for medical rehabilitation or insurance fraud detection, etc.



\section{Proposed System}

\subsection{VideoRun2D: Architecture}

\begin{figure}[t!]
    \centering
    \includegraphics[width=\columnwidth]{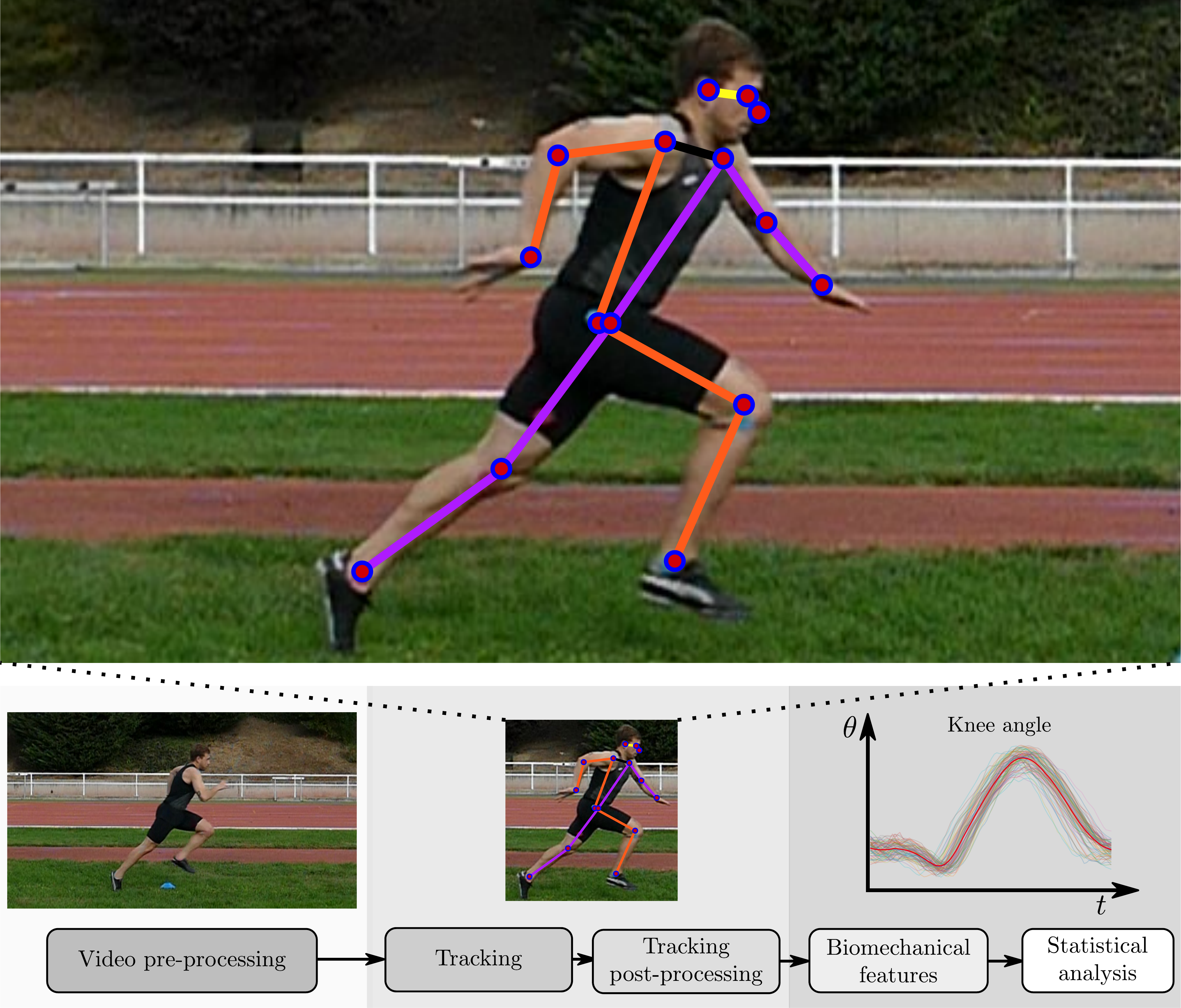}
    \caption{Block diagram of the proposed VideoRun2D system for joint angles estimation, which employs five processing modules. (In this particular example, the tracked skeleton was obtained with the MoveNet tracker core.)
    See the example video at: https://github.com/BiDAlab/VideoRun2D}
    \label{fig:BlockDiagram}
\end{figure}

Our proposed VideoRun2D performs markerless body tracking and estimates the joint angles of each user during a sprint. To achieve effective estimation, the VideoRun2D system comprises five processing modules: video pre-processing, tracking of the articular points, tracking post-processing, biomechanical features generation, and a validation system that employs statistical analysis. 
Figure~\ref{fig:BlockDiagram} shows a diagram of the proposed VideoRun2D system.

\subsection{VideoRun2D: Modules}

\subsubsection{Video pre-processing:}

This module depends both on the quality~\cite{q12fer} and characteristics of the input video and the behavior of the following tracking module. In this particular study, the input video is of very high quality~\cite{q22survey}, and the following trackers are highly robust to typical input covariates (e.g., illumination changes, typical video noise such as coding artifacts, etc.). Therefore, our only preprocessing conducted here is resizing the input videos to a resolution of 720p, enhancing processing speed. More elaborate preprocessing can be considered for VideoRun2D applications using other kind of less-quality image inputs~\cite{super19fer} or other trackers in the following module more sensible to certain images covariates~\cite{q12fer}.

\subsubsection{Tracking:}

As body tracker core in VideoRun2D, we have considered two very competitive algorithms: MoveNet and CoTracker. Both of them automatically track joint trajectories, locating the shoulder, hip, knee, and ankle joint centers of both the right and left sides.

MoveNet is a human pose estimation tool based on a bottom-up estimation model. It uses heatmaps to localize human body keypoints accurately. Google and IncludeHealth deploy it in TensorFlow Lite for modern desktops, laptops, and phones. Due to its high landmark precision, we implement the MoveNet-Thunder model in this work\footnote{https://www.kaggle.com/models/google/movenet}.

On the other hand, CoTracker is a point-tracking model proposed by Nikita~\textit{et al.}~\cite{karaev2023cotracker}. It is based on transformers~\cite{paula2023gait} and attention neural networks and tracks a dense grid of points in a video sequence. Although the CoTracker system is semi-supervised, only the initial and final points of the run will be provided to determine the trajectory in the video. CoTracker was implemented with the default parameters provided on their project page\footnote{https://co-tracker.github.io/}.

\subsubsection{Tracking post-processing:}

\begin{figure}[t!]
    \centering
    \includegraphics[width=\columnwidth]{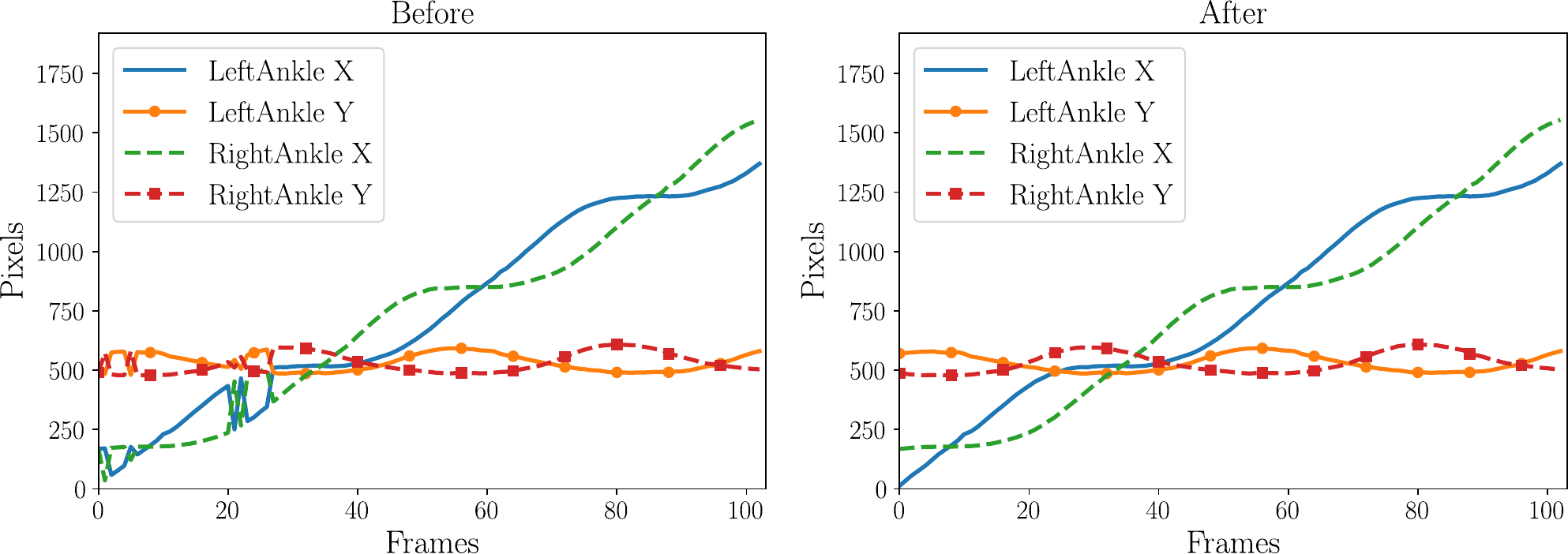}
    \caption{Graphical example of the correction of tracking errors using support vector regression, both before (on the left) and after (on the right) post-processing.}
    \label{fig:Bef_Aft_Curves}
\end{figure}

After running the tracker, three types of errors can be found due to occlusions and possible confusion in the sagittal plane. These errors are the loss of points, the confusion between the left and right side of the body, and the misallocation of a body point. To correct these errors, support vector regression was used to identify the possible outliers and then apply the necessary correction depending on the type of error. Figure~\ref{fig:Bef_Aft_Curves} shows an example of the movement of the left and right ankle before and after the applied corrections.

\subsubsection{Biomechanical features generation:}

\begin{figure}[t!]
    \centering
    \includegraphics[width=\columnwidth]{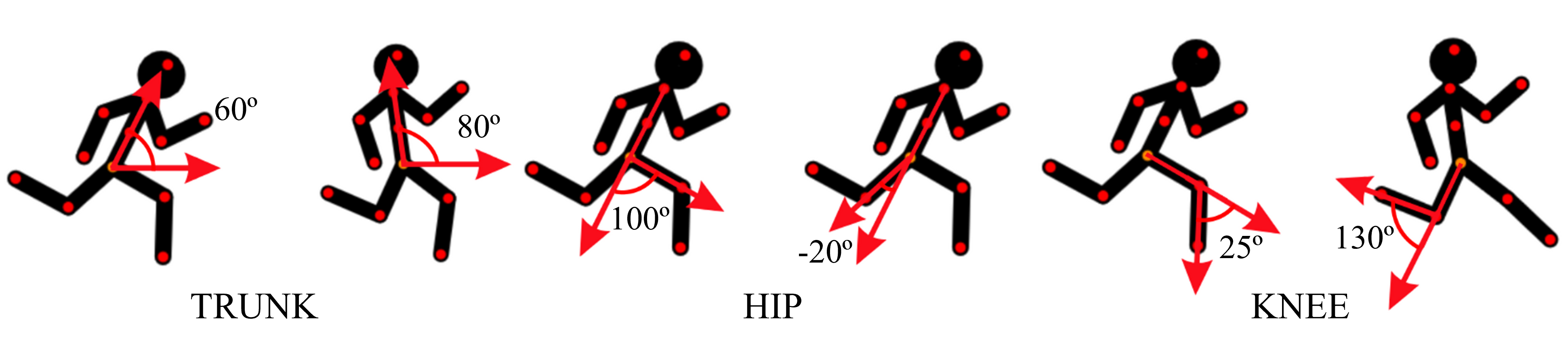}
    
    \caption{Graphical description of the joint angles variables including maximal range of motion of each joint during sprinting according to the mentioned literature on the discussion section.}
    \label{fig:JointAnglesIcon}
\end{figure}

A custom-made Python code was developed for angle estimation on the sagittal plane. The angles estimated were the inclination of the trunk relative to the horizontal, the hip flex extension (angle between trunk and thigh of both limbs), and the knee flex extension (angle between the thigh and sank of both limbs). In every angle, the positive values are associated with flexion (forward lean of the trunk) and the negative ones with extension (backward lean of the trunk). Figure~\ref{fig:JointAnglesIcon} shows three examples of angles.

\begin{figure}[t!]
    \centering
    \includegraphics[width=\columnwidth]{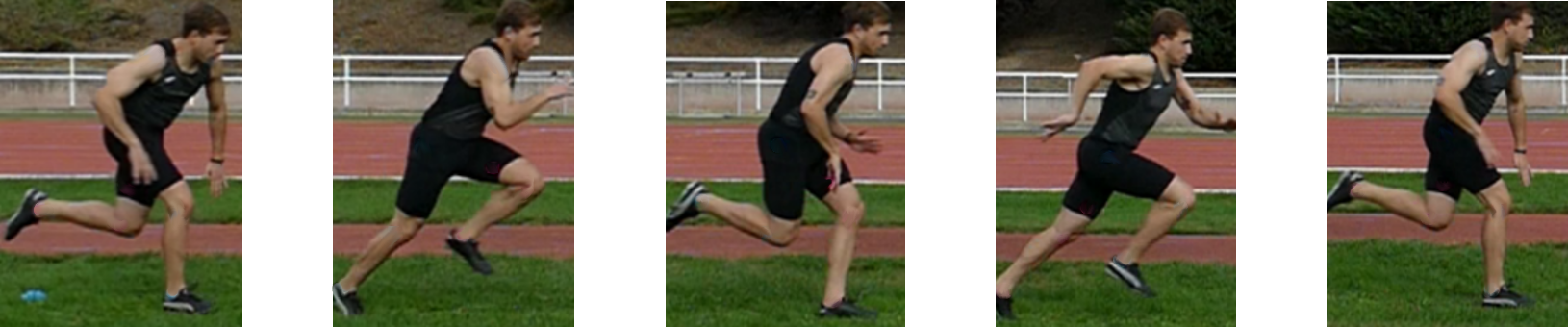}
    \caption{Example of the different stages of a stride for an individual in our dataset (which consists of 5 individuals, 40 sprints, 240 strides in total). See the example video at: \url{https://github.com/BiDAlab/VideoRun2D}}
    \label{fig:StrideExample}
\end{figure}

For each sprint, three strides of each limb are overlapped, calculating the mean of each variable. The strides are time-normalized, starting at 0\% (foot floor strike) and finishing at 100\% (next floor strike of the same foot). Each angle is calculated every 10\% of the stride. Figure~\ref{fig:StrideExample} shows the different stages of an athlete's stride.

\subsubsection{Statistical analysis}

The angles obtained from our ground truth (manually-marked Kinovea) and MoveNet are normally distributed as indicated by the homogeneous variances observed between groups. However, data from CoTracker are not normally distributed and did not pass Levene’s test. Even though, parametric statistics were used for this investigation following the statistic manual~\cite{thomas2022research}, applying the rules stabilized when the number of the samples of the groups is equal. Apart from that, for correcting the effect of multiple comparisons the Bonferroni correction was applied to the statistical tests performed.

The dependent variables analyzed are the average joint angles of the three strides in 10 normalized time intervals and the absolute errors in degrees of the markerless systems compared to the ground truth.

The independent variables studied are the three tracking systems (two markerless and one ground truth), the two types of absolute errors (ground truth compared to MoveNet and ground truth compared to CoTracker), the time intervals (each 10\% of the stride), and the subjects. One ANOVA for each angle is performed using tracking methods and time as repeated measurements (dependent variables), and subjects as independent variables. At the same time, a similar ANOVA is performed using the two errors instead of the joint angles. The level of significance was established up to $p<0.05$ and the effect size was analyzed using Generalized Squared Eta ($\eta^2$), with threshold values for small, medium, and large effects being 0.01, 0.06, and 0.14, respectively.

\section{Experiments}

\subsection{Dataset}

Five healthy amateur soccer players are analyzed (23$\pm$3.74 years old, 77.2$\pm$6.8 kg weight, 181.8$\pm$6.3 cm height). All of them signed an informed consent. The investigation was approved by the Ethics Committee of the University following the Declaration of Helsinki.

\begin{figure}[t!]
    \centering
    \includegraphics[width=\columnwidth]{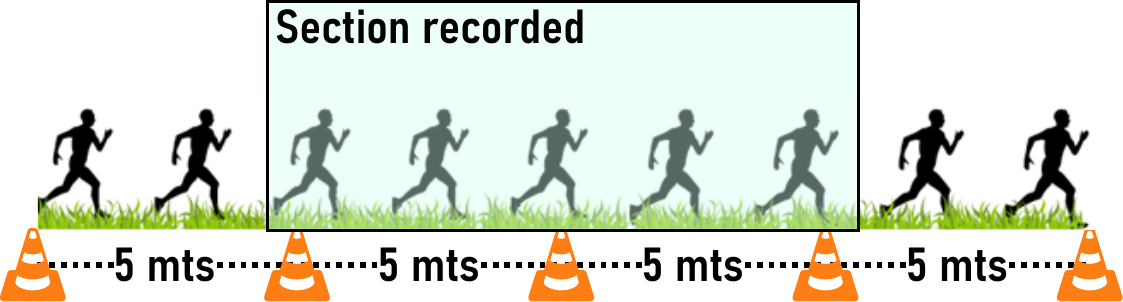}
    \caption{Data acquisition sketch.}
    \label{fig:Recording}
\end{figure}

The sprints were recorded using a video camera Panasonic DMC FZ-1000 with a resolution of $1920\times1080$p and recorded at 100 fps in two different outdoor scenarios. Each subject performed eight sprints of 20m while the camera recorded between the 5th and 15th m. The camera was placed 10 m from the middle point of the sprint. All the players sprinted with football boots on a natural grass field. Between each sprint, the players had 5 minutes of rest. A total of 40 sprints and 240 running strides were generated. The data acquisition scheme is reflected in Figure~\ref{fig:Recording}. The sprints were captured so that the optical axis of the camera was set up perpendicular to the plane of movement (sagittal plane) following the procedures of Yang and Park~\cite{yang2024improving}. See Figure~\ref{fig:StrideExample} for a visual example of the resulting recording.

\subsubsection{Ground truth based on manual labeling (Kinovea):}

\begin{figure}[t!]
    \centering
    \includegraphics[width=0.50\columnwidth]{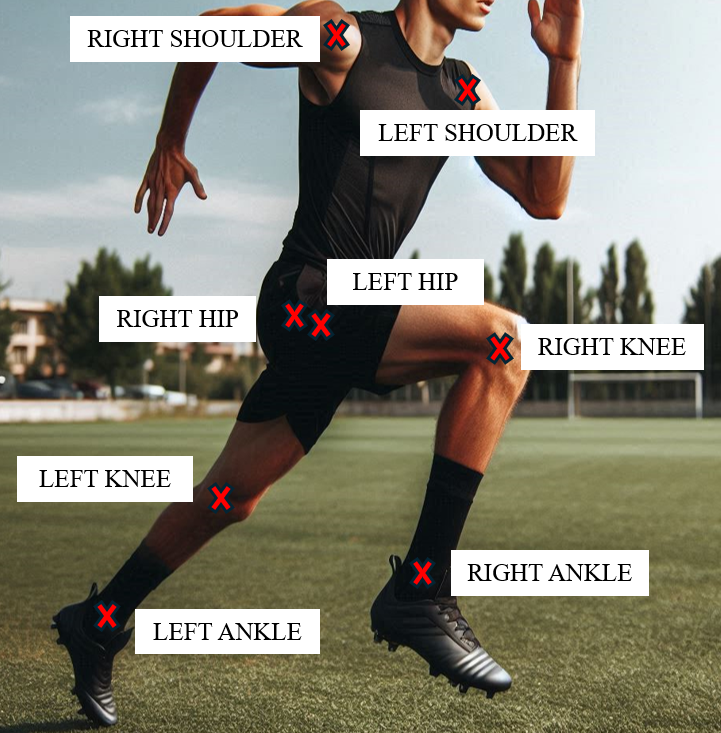}
    \caption{Location of the joint points marked on Kinovea software (ground truth).}
    \label{fig:JointPoints}
\end{figure}

Three experts previously trained manually marked all the images using the freeware Kinovea (version 0.9.5); and these marks are used as ground truth. The main instruction given to the experts was to mark the different points on the joint centers (shoulder, hip, knee, and ankle). Figure~\ref{fig:JointPoints} shows each location of the joint point marked manually. 

For the body parts that were covered by other segments, the instruction given to the experts was to mark the anatomical points as if the different body parts that covered others were transparent. These points were marked in every frame of each video. When all the frames were labeled, the trajectory of each joint was revised to find and correct possible marking mistakes.

\subsection{Results}

Tracking methods show a multivariate effect in every joint angle with a large effect size ($p \le 0.002$), meaning significant differences between them in every angle. The findings can be summarized as follows.

\begin{figure}[t!]
    \centering
    \includegraphics[width=\columnwidth]{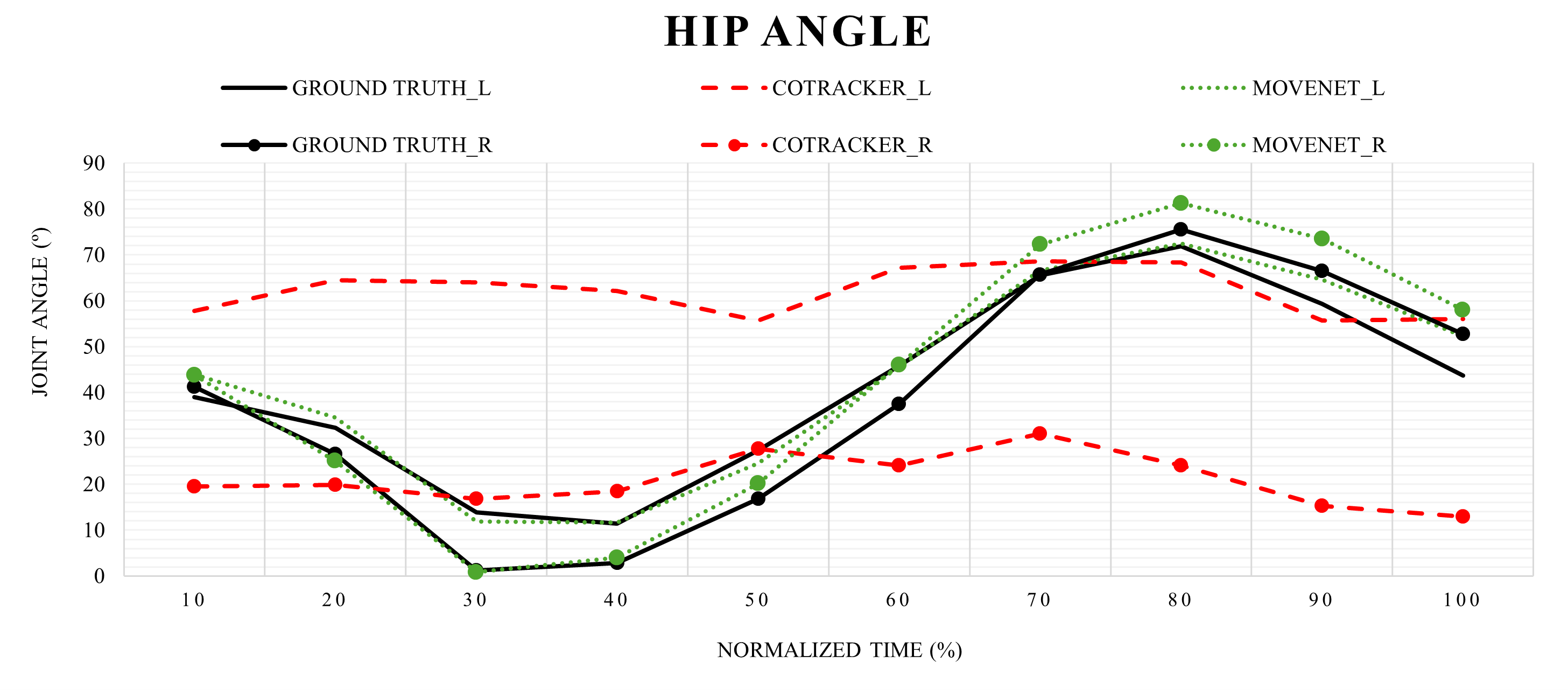}
    \caption{Flex extension pattern of the hip comparing CoTracker and MoveNet tracking methods to the manually-marked ground truth.}
    \label{fig:HipExtension}
\end{figure}

First, in the flex extension of the hip, Figure~\ref{fig:HipExtension} shows a similar behavior between ground truth and MoveNet, reaching average differences of 2° for the left hip and 4° for the right hip. A comparison between ground truth and CoTracker reveals an average difference of 20° in the left hip and 16° in the right hip.

\begin{figure}[t!]
    \centering
    \includegraphics[width=\columnwidth]{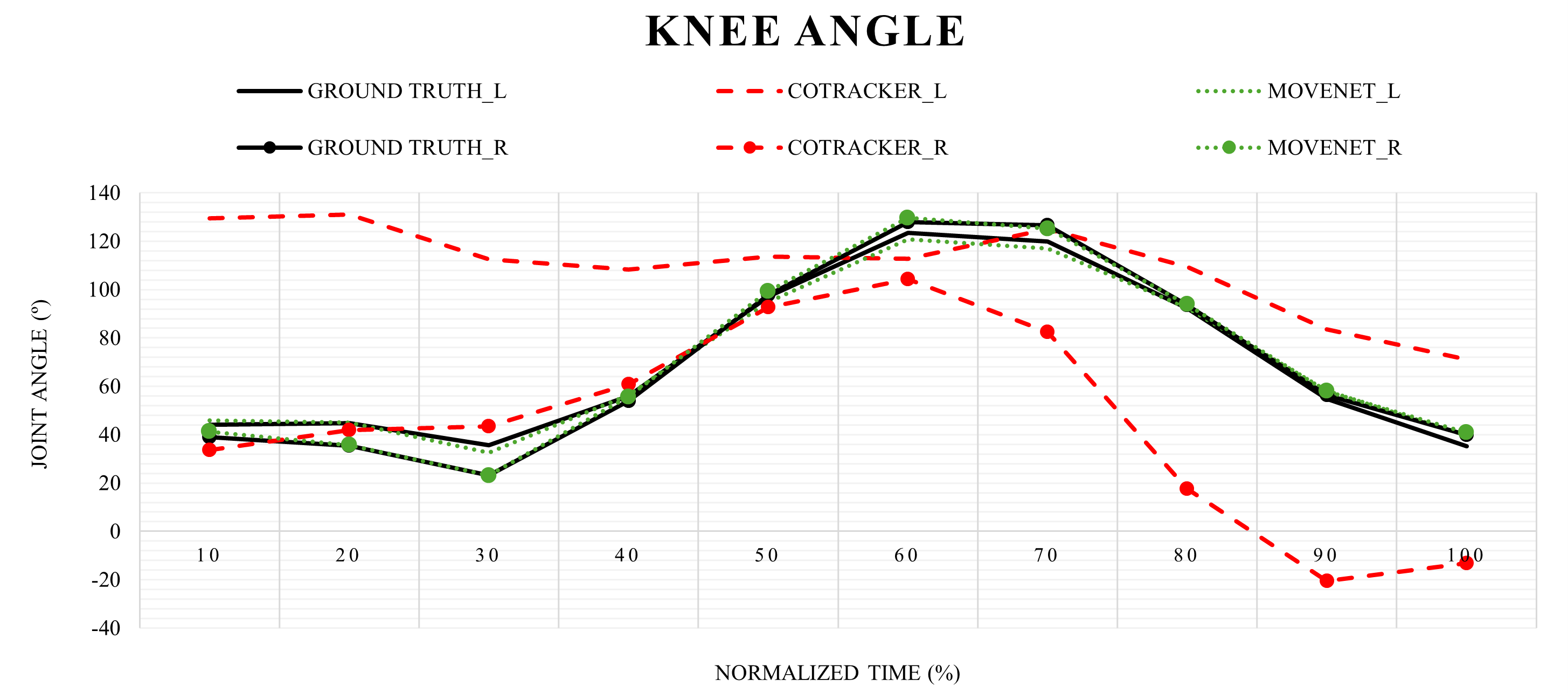}
    \caption{Flex extension pattern of the knee comparing CoTracker and MoveNet tracking methods to the manually-marked ground truth.}
    \label{fig:KneeExtension}
\end{figure}

Second, the same pattern is observed in the flex extension of the knee (Figure~\ref{fig:KneeExtension}). Ground truth and MoveNet exhibit average differences of 0.1° in the left knee and 1° in the right. Similarly, ground truth and CoTracker continue to exhibit significant discrepancies, with average differences of 39° in the left knee and 25° in the right knee.

\begin{figure}[t!]
    \centering
    \includegraphics[width=\columnwidth]{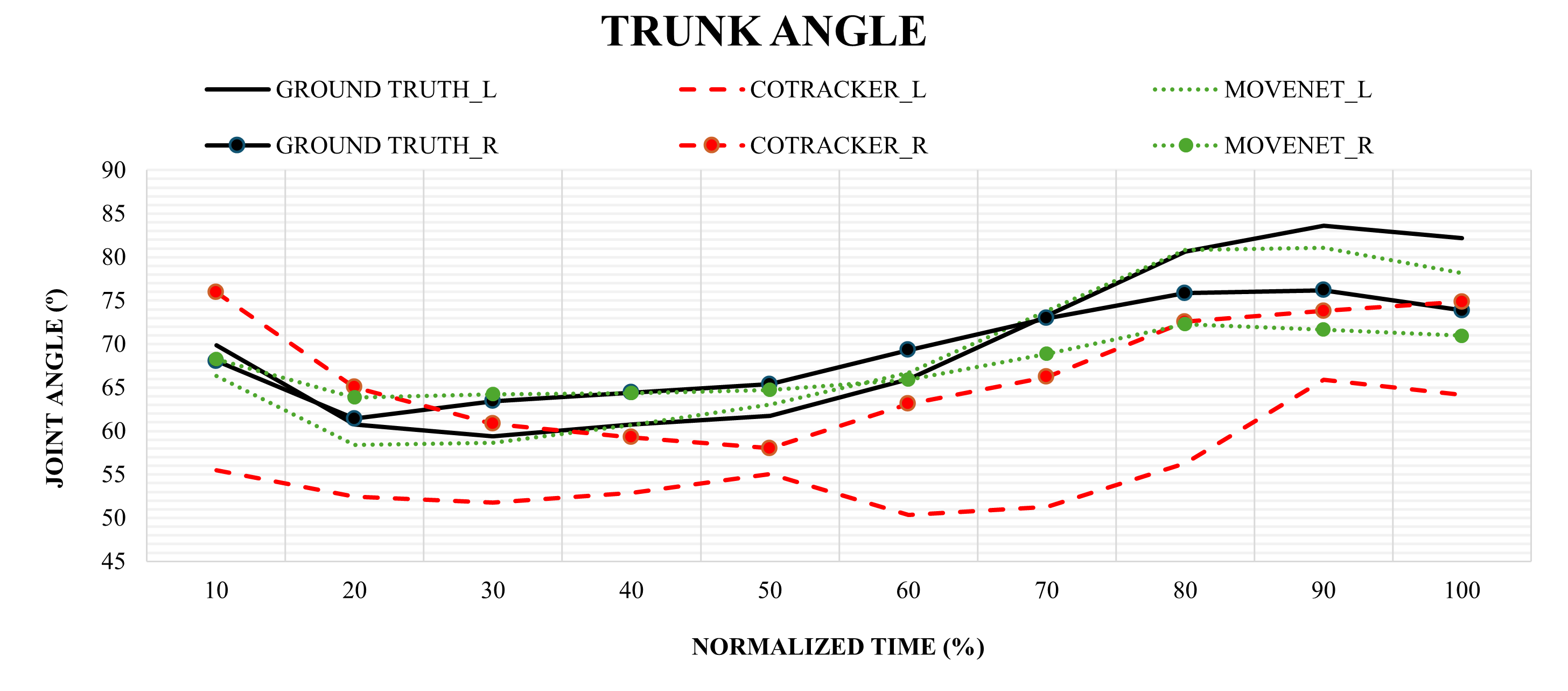}
    \caption{Flex extension pattern of the trunk comparing CoTracker and MoveNet tracking methods to the manually-marked ground truth.}
    \label{fig:TrunkExtension}
\end{figure}

Third, in the inclination of the trunk (Figure~\ref{fig:TrunkExtension}), an average difference of 1° is observed between ground truth and MoveNet on the left trunk side and 1.6° on the right side. Between CoTracker and ground truth, an average difference of 14° is observed on the left trunk side and 2° on the right side.

The tracking method-subject interaction was significant, implying that the differences vary among subjects with a large effect size ($p \le 0.001$) in the hip and knee joint angles. In contrast, no significant differences were found in the trunk angles. The tracking method-time interaction was also significant ($p \le 0.001$), implying that the differences vary over time with a large effect size in every joint angle.

The errors found in the CoTracker and MoveNet are significant ($p \le 0.001$), with a large effect size in every joint angle. The CoTracker mean error is 61.8° $\pm$ 3.6 in the left hip, 40.5° $\pm$ 3.8 in the right hip, 70.5° $\pm$ 3.3 in the left knee, 51.7° $\pm$ 4.4 in the right knee, 42.8° $\pm$ 3.8 in the left trunk, and 16.1° $\pm$ 2.0 in the right trunk. The MoveNet mean error is 5.3° $\pm$ 0.2 in the left hip, 5.5° $\pm$ 0.3 in the right hip, 3.8° $\pm$ 0.1 in the left knee, 3.2° $\pm$ 0.2 in the right knee, 3.4 $\pm$ 0.1 in the left trunk, and 3.4 $\pm$ 0.2 in the right trunk. 

The tracking method-subject-time interactions do not show significant differences in the trunk and the left hip. In contrast, significant differences in MoveNet are found in the tracking method-subject-time interaction in the knee and the right hip ($p \le 0.005$).

\begin{figure}[t!]
    \centering
    \includegraphics[width=.75\columnwidth]{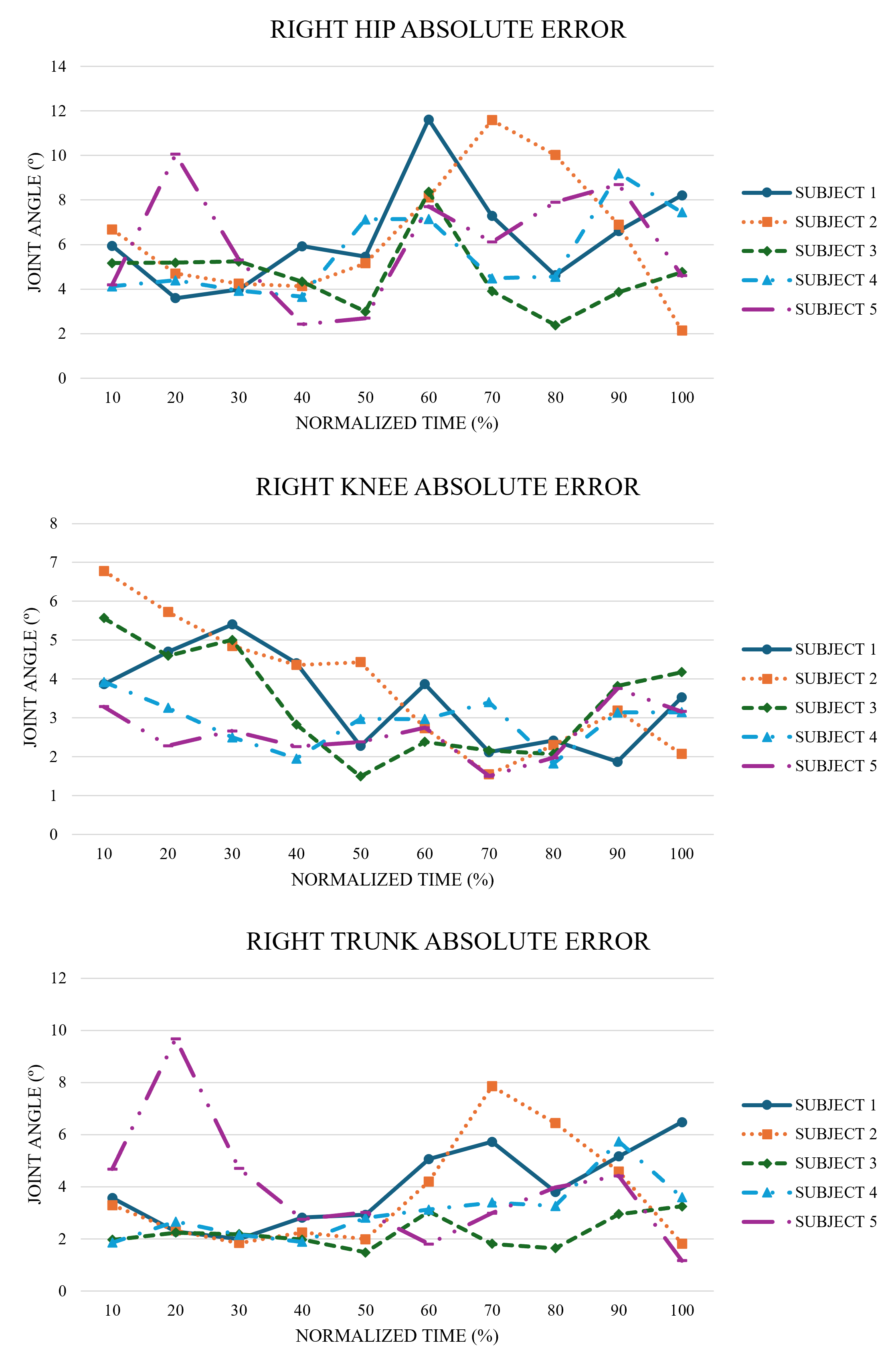}
    \caption{MoveNet absolute error of the right joints subject by subject.}
    \label{fig:AbsoluteErrorSub2SubA}
\end{figure}

\begin{figure}[t!]
    \centering
    \includegraphics[width=.75\columnwidth]{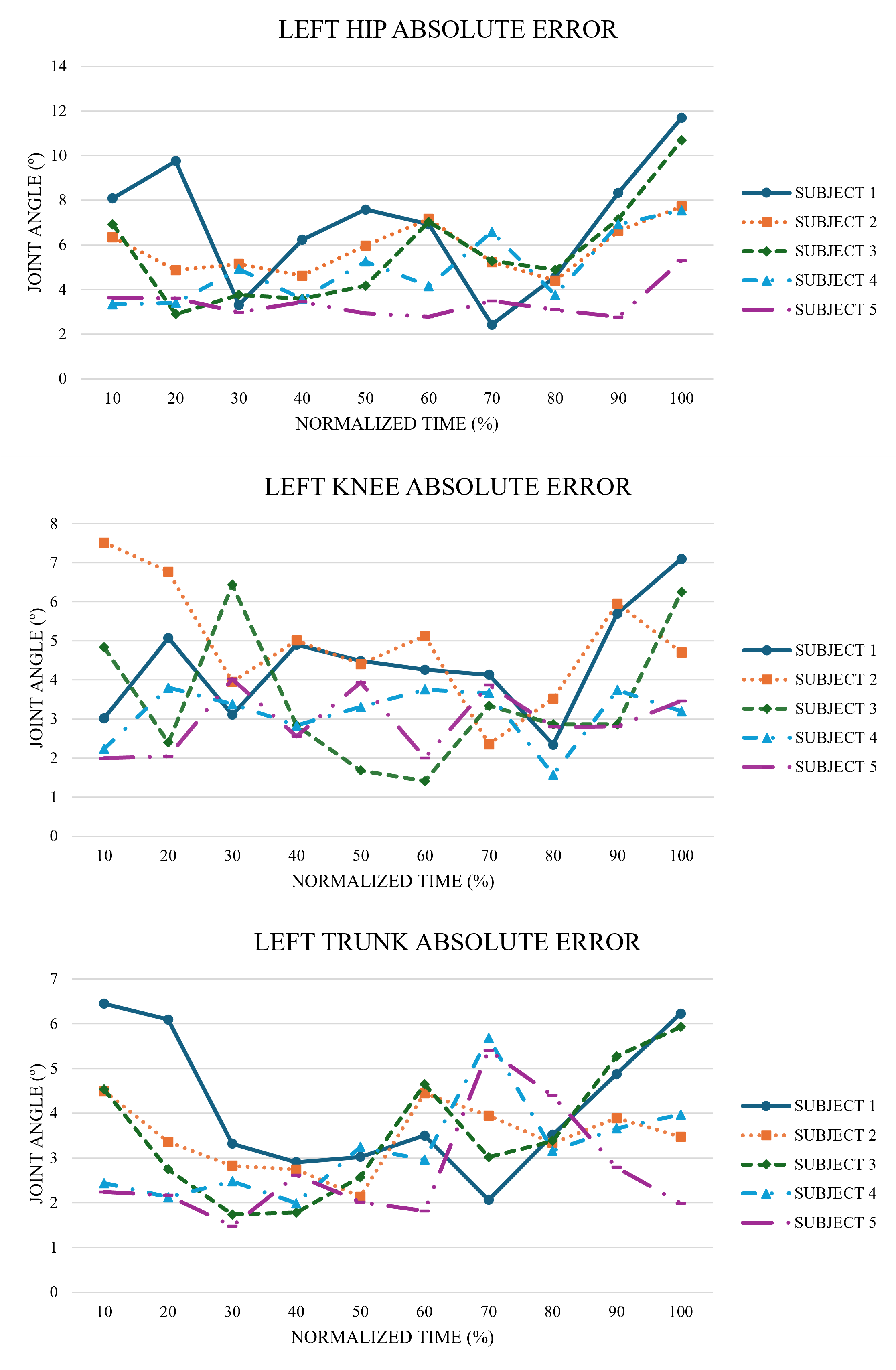}
    \caption{MoveNet absolute error of the left joints subject by subject.}
    \label{fig:AbsoluteErrorSub2SubB}
\end{figure}

A subject-by-subject analysis of the MoveNet absolute errors (Figure~\ref{fig:AbsoluteErrorSub2SubA} and Figure \ref{fig:AbsoluteErrorSub2SubB}) reveals significant differences in all joint angles exceeding 10° in the hip and 7° in the left knee for some subjects. 
In general, the error is maintained under 5°; over 5° in the right knee only in 10\% and 30\% of the stride; over 6° in the left trunk in one subject although, generally, it is under 5°; and over 8° in one subject in the right trunk side, although usually the errors remained under 6°.

\section{Discussion and Conclusion}

This study evaluated two trackers (CoTracker and MoveNet) against manual labeling for sprint analysis. We adapted and wrapped those trackers for this particular application area, including proper pre- and post-processing, generating the markerless motion capture system termed VideoRun2D.
%
%
In our experiments, we have evaluated VideoRun2D using biomechanical analysis of joint angle kinematics in a dataset of 40 sprints videos from 5 athletes.\footnote{Example videos and the full dataset generated for our experiments are available here: https://github.com/BiDAlab/VideoRun2D}
Manual labeling was considered the ground truth, and the VideoRun2D errors with respect to the ground truth were calculated. MoveNet could estimate the flex extension patterns well in every joint angle, similar to the ground truth. 
This conclusion agrees with Ota~\textit{et al.} in 2021, who evaluated the joint kinematics with a marker-based system and a markerless one, showing that the markerless one followed the same pattern as the marker-based system~\cite{ota2021verification}.

In contrast, the CoTracker method did not perform a good estimation, according to Cronin~\textit{et al.}, who concluded that some markerless systems are not suitable for accurately tracking an athletic movement~\cite{cronin2024feasibility}.

The CoTracker method yielded an average difference between 2° and 39° compared to ground truth, indicating that it is not an appropriate method for estimating joint angles. On the contrary, the MoveNet method demonstrated a lower average difference,  from 0.1° to 4°, thereby exhibiting a better precision. When observing each joint separately, the knee flex extension was where the MoveNet method showed better results (0.1°-1° average differences) compared to the ground truth, followed by the trunk (1°-1.6°) and the hip (2°-4°).

These average differences should be interpreted cautiously as differences that overestimate or underestimate the joint angles offset one another, reducing the degrees of average differences.

When analyzing the curves of each joint, the pattern of the hip is the same that was documented by Lee~\textit{et al.} in 2009, Lin~\textit{et al.} in 2023, and Okudaria~\textit{et al.} in 2022, who observed the sprint with a top-performing marker-based motion capture system~\cite{lee2009running,lin2023validity,okudaira2021sprinting}. However, Lin~\textit{et al.} found a higher extension of the hip reaching until -20°~\cite{lin2023validity}, and Okudaria~\textit{et al.} analyzed sprints with different inclinations, observing the same pattern as our methods but finding a higher flexion of the hip (100°)~\cite{okudaira2021sprinting}.
The knee pattern is similar to the curves obtained in the references from a marker-based motion capture system during sprinting and during sprinting with different inclinations~\cite{lee2009running,lin2023validity,okudaira2021sprinting}.

The trunk curve must be clarified, as the range of motion is less than in other joints. The pattern obtained agrees with the one studied by Nagahara~\textit{et al.}, who studied the behavior of the trunk in different parts of the sprint, concluding that when accelerating, the trunk angle is more horizontal. When the velocity rises, the trunk is straightened~\cite{nagahara2017kinematics}. Their results showed that the trunk angle tilts forward during flight time and tilts backward during the support phase~\cite{nagahara2017kinematics}.

MoveNet errors oscillate between 3.2° and 5.5°, smaller than errors from other markerless systems used by other authors. Viswakumar~\textit{et al.} 2022 found errors in a markerless system of 7.73° in the hip and 5.82° in the knee flex extension~\cite{viswakumar2022development}. Also, they demonstrated that markerless systems have higher accuracy than Kinect cameras, as the markerless system adapts to different clothes and light situations better~\cite{viswakumar2022development}. Other authors, such as Yang and Park in 2024 and Stenum~\textit{et al.} in 2021, reported markerless errors similar to the ones calculated in this investigation (between 4° y 5.6°)~\cite{stenum2021two,yang2024improving}.

When comparing markerless systems with other methods, IMUs showed errors under 10°~\cite{lin2023validity}, indicating that our method works better without placing any device over the athlete’s skin.

The results did not show errors’ consistency among subjects or time, making it harder to correct them. When watching the data subject by subject, the errors in some participants increased, reaching 10° in the hip, 5°-7° in the knee, and 6°-8° in the trunk. These results agree with the ones obtained by Cronin~\textit{et al.} in 2023, which shows how the markerless system angle estimation varies among subjects~\cite{cronin2024feasibility}. Additionally, the errors varied significantly across the stride time, increasing in the first part and at the end. Therefore, we are working now in applying user-dependent~\cite{ud05-3,ud05-1,ud05-2} and time-adaptive methods~\cite{fusion18,time18} to improve VideoRun2D for future versions.

Other studies have investigated markerless systems, concluding that they are not an alternative to marker-based systems due to errors in the front plane or calculating other variables rather than sagittal angles~\cite{cronin2024feasibility,ota2021verification}. Even when analyzing sagittal angles, there is a lack of precision in certain moments of the stride.

In conclusion, we have found that CoTracker is not viable as core tracker for evaluating sprint kinematics in our proposed VideoRun2D markerless motion capture system. On the other hand, MoveNet has the potential to detect a similar pattern to marker-based ones of joint kinematics during a sprint. 

At the present time, the precision of VideoRun2D is good but still improvable in comparison to much more expensive marker-based solutions. Towards that improvement we will work in better postprocessing to reduce the errors relative to manual labeling, regulate the errors among subjects, and understand why the errors increase, particularly at the beginning and end of the stride.

\begin{credits}
\subsubsection{\ackname} Project BBforTAI (PID2021-127641OB-I00 MICINN/FEDER). A. Morales is also supported by the Madrid Government under the Multiannual Agreement with UAM in the line of Excellence for the University Teaching Staff (V PRICIT). G. Garrido-López is also supported by Banco Santander under a doctorate grant.

\end{credits}
%
%
%
%
\bibliographystyle{abbrv}
\bibliography{main}

\end{document}